\documentclass[letterpaper, 10 pt, conference]{ieeeconf}  % Comment this line out if you need a4paper

\IEEEoverridecommandlockouts                              % This command is only needed if 
\usepackage{graphics} % for pdf, bitmapped graphics files
\usepackage{epsfig} % for postscript graphics files
\usepackage{mathptmx} % assumes new font selection scheme installed
\usepackage{times} % assumes new font selection scheme installed
\usepackage{amsmath} % assumes amsmath package installed
\usepackage{amssymb}  % assumes amsmath package installed

\usepackage{algorithm}
\usepackage{algorithmic}
\usepackage{subcaption}
\usepackage{float}
\usepackage{caption}
\usepackage{url}
\usepackage{hyperref}
\usepackage{amsmath}
\usepackage{booktabs} 
\usepackage{multirow}
\usepackage{graphicx}
\usepackage{float}  % just in case, not required here
\usepackage{newfloat}
\usepackage{listings}
\usepackage{tabularx,makecell,array}
\newcolumntype{Y}{>{\centering\arraybackslash}X}

\usepackage{amsmath}

\DeclareMathOperator*{\argmin}{arg\,min}

\usepackage{tabularx,makecell,array}
\usepackage{url}
\usepackage{titlesec}

\title{\LARGE \bf
Latent Representations for Visual Proprioception in Inexpensive Robots}

\author{Sahara Sheikholeslami and Ladislau Bölöni\\
University of Central Florida\\
4000 Central Florida Av., Orlando FL 32816\\
{\tt\small sahar.sheikholeslami@ucf.edu, ladislau.boloni@ucf.edu}
}

\begin{document}

\maketitle
\thispagestyle{empty}
\pagestyle{empty}

\begin{abstract}

Robotic manipulation requires explicit or implicit knowledge of the robot's joint positions. Precise proprioception is standard in high-quality industrial robots but is often unavailable in inexpensive robots operating in unstructured environments. In this paper, we ask: to what extent can a fast, single-pass regression architecture perform visual proprioception from a single external camera image, available even in the simplest manipulation settings? We explore several latent representations, including CNNs, VAEs, ViTs, and bags of uncalibrated fiducial markers, using fine-tuning techniques adapted to the limited data available. We evaluate the achievable accuracy through experiments on an inexpensive 6-DoF robot.

\end{abstract}    
\section{Introduction}

{\em Proprioception} is the task of recovering the configuration of the robot from its own sensors, in contrast to {\em perception}, which is directed towards the external reality. In some settings, proprioception is an engineering problem solved by the internal sensors of the robot. For instance, high-quality industrial robots are so precisely actuated that we can safely consider their joint configurations known. 

However, for certain scenarios, such as inexpensive robots operating in unstructured environments, the proprioception information coming from the robot might be noisy, uncertain, or unreliable. These robots might be controlled through policies based on end-to-end reinforcement learning or imitation learning that define actions as functions of an external observation $a \leftarrow \pi(o)$, which appears to sidestep the proprioception problem. In practice, however, if some internal proprioception is available, this can be combined with the results of the external perception, in the hope that explicit proprioceptive data can support task performance.

\begin{figure}
    \centering
    \includegraphics[width=0.6\linewidth]{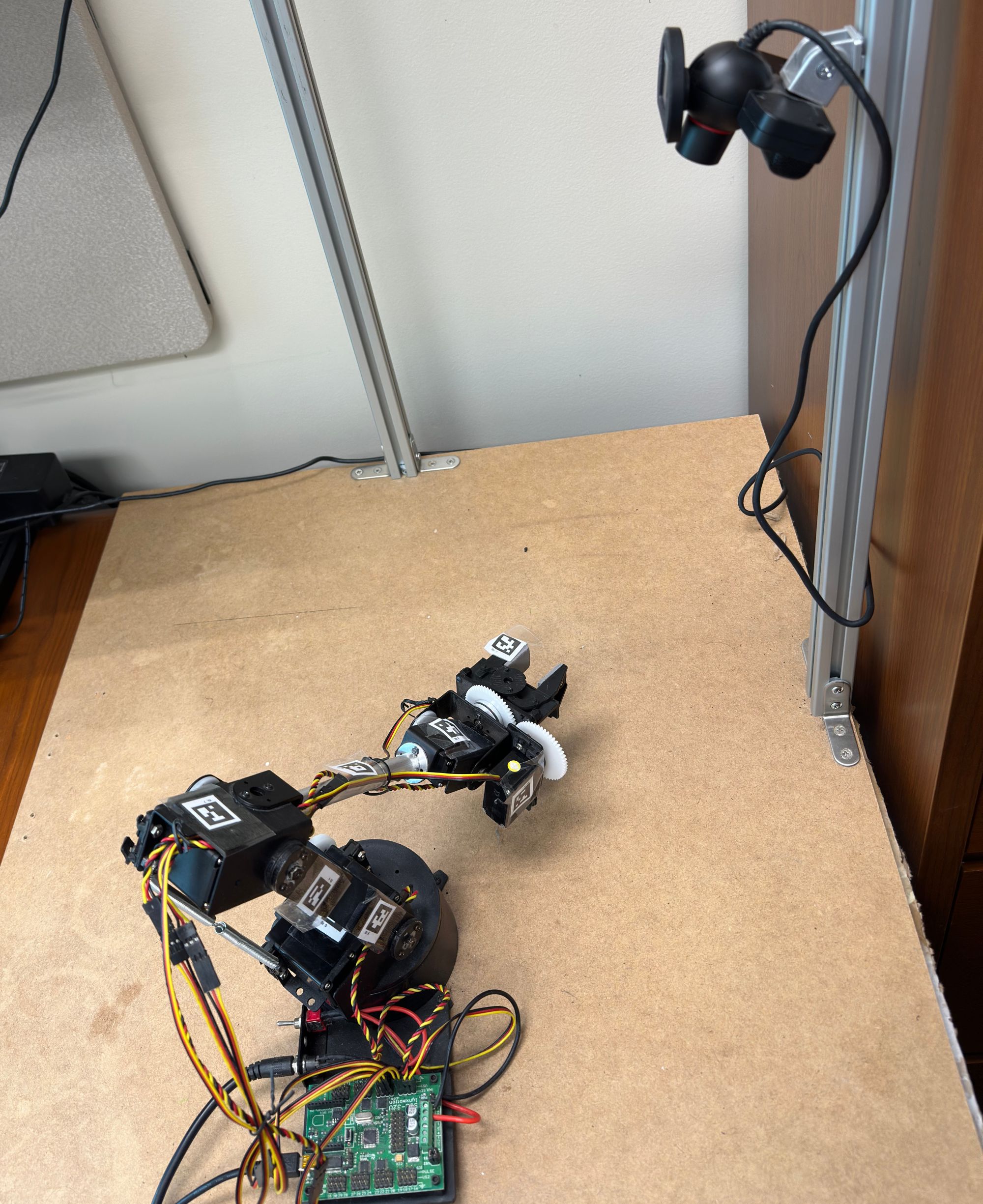}
    \caption{The experimental setup for visual proprioception. An inexpensive robot with six degrees of freedom (Lynxmotion AL5D) is being observed by a low-resolution camera. Neither the robot nor the camera is calibrated. The objective is to recover the configuration of the robot from a single captured RGB image.}
    \label{fig:RobotSetup}
\end{figure}

\begin{figure}
    \begin{subfigure}[b]{0.45\linewidth}
        \centering
        \includegraphics[width=\linewidth]{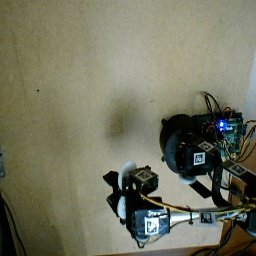}
    \end{subfigure}
    \hfill
    \begin{subfigure}[b]{0.45\linewidth}
        \centering
        \includegraphics[width=\linewidth]{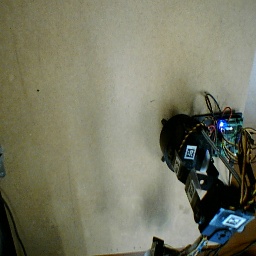}
    \end{subfigure}
    \vspace{2mm}

    \begin{subfigure}[b]{0.45\linewidth}
        \centering
        \includegraphics[width=\linewidth]{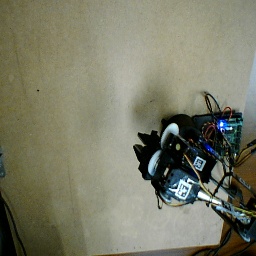}
    \end{subfigure}
    \hfill
    \begin{subfigure}[b]{0.45\linewidth}
        \centering
        \includegraphics[width=\linewidth]{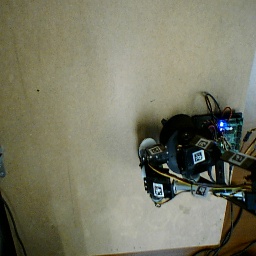}
    \end{subfigure}
    \caption{Four robot observations illustrating visual proprioception challenges: gripper state is visible in the left images but occluded or out of frame on the right, and viewing angle makes height estimation difficult even for humans.}
    \label{fig:pictures}
\end{figure}
In this paper, we ask the question: Can an inexpensive robot recover some part of its configuration, through a fast and architecturally straightforward visual proprioception technique? We expect this information to be used not as ground truth in a classical control model, but as one of the many noisy inputs to a learned policy. It is not our objective to push the boundary of accuracy, which would require more and better cameras, more compute and sophisticated, possibly iterative, optimization techniques. Instead, we aim to push the boundary of low cost along several dimensions. Which technique requires the least computation, handles lower-quality sensor input, and has lower requirements with regard to the training data and access to physical and visual models?

% However, proprioception remains essential in many scenarios. Several classes of robots lack actuation accuracy sufficient to neglect proprioception at any level of abstraction. Low-cost robots may have limited sensing capabilities, while underactuated, compliant, or mobile-mounted systems often require proprioceptive feedback. Damaged robots also rely on proprioception to compensate for structural impairments. Additionally, proprioception aids in recovering a robot's configuration when displaced by external forces. A notable example is kinesthetic teaching, where a human physically guides the robot to demonstrate a task. Furthermore, when a robot manipulates an external tool, such as a spoon or scissors, the tool effectively becomes an extension of the robot without the same level of instrumentation, necessitating proprioceptive estimation.

% This paper proposes and evaluates several end-to-end learned approaches to {\em visual proprioception} for robot manipulators. 

Our setup involves a robot arm observed through a fixed-position, uncalibrated camera, using the captured images to estimate its configuration (see Fig.~\ref{fig:RobotSetup}). This approach is cost-effective and commonly used in many installations. The camera feed remains available for other control tasks. While we do not claim biological plausibility for any of the algorithms proposed, it is noteworthy that most human manipulation occurs under a similar arrangement -- observing one's own hands from an external viewpoint. Our choice of such a simple configuration is intentional, as this setting has been significantly underexplored in the literature. We do not consider multiple cameras, calibrated cameras, depth cameras, wrist mounted cameras, proprioception using a video stream, explicit access to geometric constraints, kinematic models of the robot, access to a simulator, ability to render and compare the pose of the robot and information from robot sensors. As the literature shows, any of these can improve the quality of proprioception. Nevertheless, we argue that our resource-deprived setting is a worthy subject of study, because the types of algorithms that succeed in this setting have to be very different from those that are appropriate in other scenarios.

Within this setup, we assume a minimalistic, single-pass architecture for visual proprioception. A computer vision algorithm generates a modest-sized \textit{proprioception-dedicated latent representation} $\mathbf{z}_\mathit{prop}$ from a single image, and an MLP regressor extracts proprioceptive information from this representation. The only information available about the robot architecture is \textit{some} supervised training data in the form of known image/configuration pairs -- sufficient to support fine-tuning, but not enough to train models from scratch. Additionally, we assume access to the robot, allowing the collection of unsupervised training data in the form of images of the robot in various, not necessarily known, configurations.

In this architecture, the key determinant of performance is the latent representation $\mathbf{z}_\mathit{prop}$. Given that the robot configuration is low-dimensional (six degrees of freedom in our case), we hypothesize that a latent representation of modest dimensionality—128 or 256—should suffice. As for the algorithm used to compute the latent encoding, we do not exclude any classical or modern computer vision methods, as long as they operate within the constraints of the proposed framework.

The main contributions of this paper are as follows:

\begin{itemize}
% [left=0pt, labelindent=0pt, itemindent=0pt, labelsep=0.5em, itemsep=0pt]
    \item We propose four alternative techniques for creating a latent representation suitable for visual proprioception. These are based on convolutional variational autoencoders, pre-trained vision backbones (convolutional neural nets, and vision transformers) with reduced latent spaces fine-tuned for proprioception and "bags" of randomly placed, uncalibrated fiducial markers. 
    
    \item We introduce an architecture for extracting proprioception information that can be used across different latent representations and sizes without structural modifications, requiring only retraining on a minimal amount of supervised data.
    
    \item We conduct experiments to compare the performance of the proposed representations, including two latent-size variants. Results show variations in error rates across configuration components, with some representations better suited for specific measures. Different representations exhibit distinct error and noise patterns, providing insights relevant to both robot design and downstream applications that utilize proprioception data.
\end{itemize}

\section{Related work}
\label{sec:RelatedWork}

\noindent {\bf Proprioception from visual input} has been studied in the robotics literature for a number of different scenarios.
% IROS-2016, 16 citations
Ortenzi et al.~\cite{ortenzi2016vision} implement vision-based proprioception motivated by applications such as nuclear decommissioning, where electronic sensors might be vulnerable to radiation. Their approach is based on tracking several parts of the robot from a monocular camera and optimizing a set of transformation matrices that relate the camera, world, and tracked robot parts. In a follow-up work~\cite{ortenzi2018vision}, the accuracy was improved by adding custom-designed fiducial markers to the joints. 
% ICRA 2016
Widmaier et al.~\cite{Widmaier-2016-ArmPoseEstimation} propose a technique for arm pose estimation, motivated by the errors in the joint encoders or inaccurate calibration. The approach is based on applying a random regression forest to pixel-by-pixel features calculated from segmented depth images, and operates on a frame-by-frame basis, with no correlation expected between timesteps. 

% Well cited paper, inexpensive robot toy
Zuo et al.~\cite{Zuo-2019-Craves} describe CRAVES, a system for performing visual proprioception on an inexpensive robot that lacks sensors. The approach is based on first detecting 2D keypoints on the robot using a two-stack hourglass network, and then calculating the pose by recovering the 3D pose under an assumed perspective projection. The model is trained on synthetic data generated from an Unreal Engine-based simulator. In its minimal deployment requirements, CRAVES is the closest work to ours. For training, however, CRAVES assumes the existence of a simulator capable of visually rendering various poses, which is not required in our approach. 

Rauch et al.~\cite{rauch2019learning} propose a tracking framework for robotic manipulators using RGB-D images, where a rough estimate is first provided via depth keypoints, which are then refined through information about the color edges. 
% IROS-2020, only 6 citations
% In this paper, they are estimating the position of the end-effector from a depth sensor. 
% !!! End-effector only
Cheng et al.~\cite{cheng2020real} estimate the position of the end-effector of a Kinova Jaco 2 robot from the depth images collected using a Kinect device. 
Liu et al.~\cite{liu2020nothing} introduce GC-Pose, an unsupervised and model-free method that estimates joint configurations of a robotic arm (or other articulated objects) using only RGB or RGB-D images. 
% CVPR-2021, 50 citations
Labbé et al.~\cite{labbe2021robopose} introduce RoboPose, a render-and-compare framework for markerless 6D pose and joint angle estimation from a single RGB image. The approach assumes the existence of a known model of the robot and an ability to create a visual rendering of the robot. Trained on synthetic data, it iteratively refines predictions, overcoming self-occlusion and depth ambiguity for higher accuracy than keypoint methods.

Simoni et al.~\cite{Simoni2022} propose a 3D robot pose estimation framework using depth cameras. The approach introduces a new intermediate representation, called semi-perspective decoupled heatmaps, from which the 3D joint positions are calculated. 
% ECCV 2024, 5 citations. They are using custom networks for the various components of the configuration
Ban et al.~\cite{Ban2024} introduce a real-time framework for holistic robot pose estimation from a single RGB image, predicting 6D pose and joint states in one pass. The approach trains dedicated neural networks for subtasks such as estimation of camera-to-robot rotation, robot-joint states, or robot root relative keypoint locations. Tian et al.~\cite{Tian2024} propose RoboKeyGen, a keypoint-based approach that decouples the joint prediction problem into two sub-tasks: the detection of keypoints in a 2D image, and the lifting of these 2D keypoints into 3D by a perspective transformation that takes into consideration the robot's structural information. The joint angles are then regressed from these 3D keypoints. 

\noindent {\bf Camera-pose calibration:} A closely related task to visual proprioception is determining the pose of the {\em camera} relative to the robot's root (or possibly the gripper), often in the context of known joint configurations. The two main approaches found in the literature are those based on keypoint detection and matching and the techniques based on rendering the robot. 
% cited by 131 ICRA-2020, Dieter Fox
Lee et al.~\cite{Lee-2020-ICRA} base their method on detecting keypoints (such as joints) associated with the robot using a deep neural network trained on simulated data with domain randomization. With the assumption that the joint configuration is known, the camera position can be recovered using Perspective-n-Point algorithms. 
% cited by 20, CVPR-2023
Lu, Richter and Yip~\cite{Lu-2023-CVPR} solve camera pose determination by a rendering-based approach using a fully differentiable renderer, allowing the system to match the speed of typical keypoint-based approaches. 
Han et al.~\cite{Han2024} describe PoseFusion, a keypoint-based framework for markerless camera-to-robot pose estimation, which uses nested U-structures that capture contextual information across various scales. 

\noindent {\bf Other visual proprioception scenarios:} Estimating robot state from visual information appears in many other application scenarios that are more distantly related to our setup. Soft robots, for instance, often implement proprioception with an {\em internal} vision-based sensor~\cite{Zhang-2023-SoftRobotProprioception}. Cong et al.~\cite{cong2022} propose a vision-proprioception RL model for robotic planar pushing. Building on this work, ~\cite{bergmann2024precision} increase precision by adding a memory feature.

%%% FIXME more can be added here

% This is not really doing proprioception. 
 % Marker-free and superior to raw image policies, it generalizes well in real-world tests.

\section{Representations for visual proprioception}
\label{sec:Encodings}

\subsection{Preliminaries}
\label{sec:Preliminaries}

Let us consider a robot manipulator whose configuration is described by a six-dimensional vector $\mathbf{a} = [a_1,\dots, a_6]$ with $a_1$ the {\em height} of the gripper, $a_2$ the {\em distance} of the gripper from the robot base, $a_3$ the {\em heading} of the robot arm, $a_4$ the up-down {\em angle of the wrist}, $a_5$ the {\em rotation of the wrist} around the arm, and $a_6$ the {\em gripper status} between open and closed, with all values normalized to $[0,1]$. We search for a parameterized function $f(o; \theta) \rightarrow \hat{\mathbf{a}}$, where the observation $o$ is an RGB image input and $\hat{\mathbf{a}}$ is an estimate of the robot configuration. 

Figure~\ref{fig:pictures} illustrates the challenges of this task. Even for a human observer, it is difficult to assess the height of the gripper due to the viewing angle. While the gripper's open/closed status is clearly visible in the top-left and bottom-left pictures, it is either outside the image or obstructed by other parts of the robot in the images on the right-hand side. The gripper assembly is completely out of view in the top-right image.

% FIXME: many other citations here
Deep learning-based robot control usually relies on a latent representation $\mathbf{z} = h(o)$, sometimes present in an explicit form~\cite{Rahmatizadeh-2018-InexpensiveRobot}, while in other cases, for instance in vision-language-action models, implicitly~\cite{Kim-OpenVLA}. Learned representations often have no interpretable structure, although some approaches aim to create structure through training~\cite{Abolghasemi-2020-ICRA}. To separate the concerns of proprioception and control, we consider a compact latent representation {\em dedicated to proprioception} $\mathbf{z}_\textit{prop}$ (which might be part of a larger latent space $\mathbf{z}_\textit{full}$). To avoid notation clutter, in the rest of this paper $\mathbf{z}$ will refer to $\mathbf{z}_\textit{prop}$, and, correspondingly, $h$ will refer to proprioception-dedicated sensor processing $\mathbf{z}_\textit{prop} = h_\textit{prop}(o)$.

Starting from a dataset of pairs of observation and configuration ground truth: $\mathcal{D} = \{ (o_1, \mathbf{a}_1), \ldots (o_n, \mathbf{a}_n) \}$ and an encoder $\mathbf{z} = h(o)$ the proprioception regression problem can be defined as 
\begin{equation}
    \theta^* = \argmin_\theta \frac{1}{n} \sum_i dist(\mathbf{a}_i, f(h(o), \theta))\,.
\end{equation}
Following standard practices, we implement the regression function $f(\cdot)$ as a multilayer perceptron with an input layer matching the dimensionality of $\mathbf{z}$ (128 or 256), two hidden layers each with a dimensionality of 64, and an output layer of size 6, matching the dimensionality of the configuration space $\mathbf{a}$ (see Fig~\ref{fig:ProprioceptionRegression}). This architecture is agnostic to the latent representation; naturally, for each representation, the network was retrained from scratch, with the trained weights becoming part of the final proprioception architecture. We used early stopping based on validation loss, and found that the training was fast and stable for each latent representation. 

In addition, the same regression architecture was used to provide a surrogate loss in the fine-tuning of the CNN- and ViT-based latent representations. However, the weights trained for the surrogate loss were discarded after fine-tuning. 

\subsection{Latent encoder alternatives for proprioception}

The latent encoder is the most critical part of a visual proprioception model, with a wide range of possible architectures ranging from classical computer vision techniques based on handcrafted features to visual transformers and state space models. In line with our focus on inexpensive robots controlled in real-time, we are focusing on architectures where the latent extraction $h(o)$ can run at a granularity of tens of milliseconds on an {\em edge compute} device deployed in the field. This excludes from consideration, for instance, large language models with multi-step reasoning, or expensive segmentation models. We note, however, that this consideration only applies to the inference of the visual proprioception latent. We do not expect the model training to be performed on edge devices, and it is perfectly possible for the high level control of the robot, which requires a granularity of several seconds, to be performed using computationally demanding models, such as VLAs. 

In the following, we briefly discuss the set of alternative latent encoders we implemented and compared.

\newcommand{\mywidth}{0.30\linewidth}
\newcommand{\myheight}{3.4cm}
\begin{figure*}
    \centering
    \begin{subfigure}[b]{0.32\linewidth}
        \centering        \includegraphics[height=\myheight]{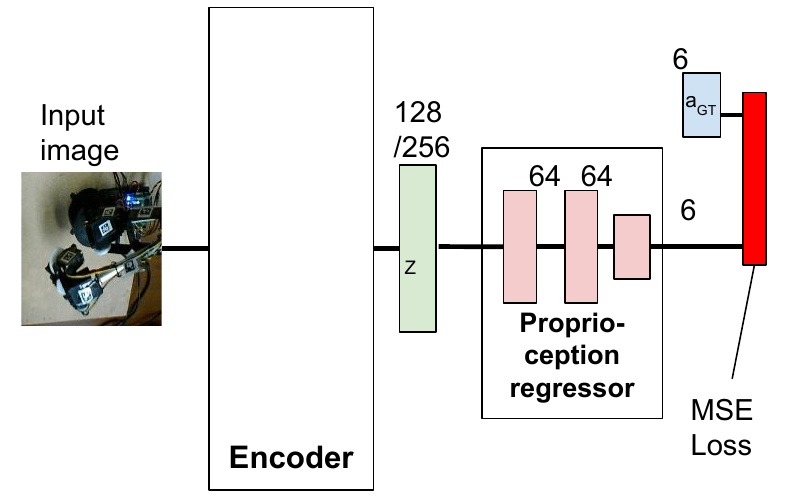}
        \caption{Proprioception regression from a latent representation.}
        \label{fig:ProprioceptionRegression}
    \end{subfigure}
    \hspace{3mm}
    \begin{subfigure}[b]{\mywidth}
        \centering
        \includegraphics[height=\myheight]{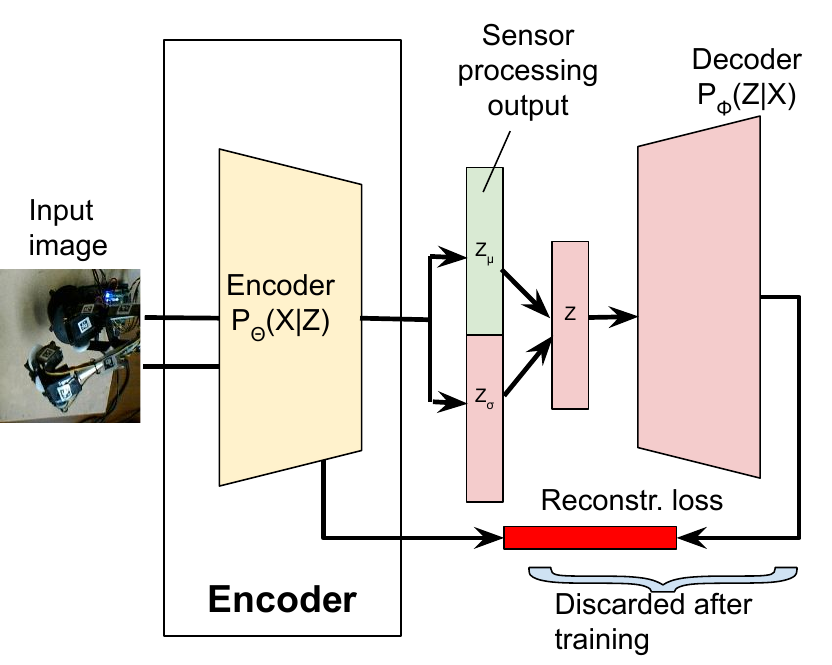}
        \caption{Encoder based on a convolutional variational autoencoder.}
        \label{fig:encoding-convvae}
    \end{subfigure}
    \hspace{3mm}
    \begin{subfigure}[b]{0.32\linewidth}
        \centering
        \includegraphics[height=\myheight]{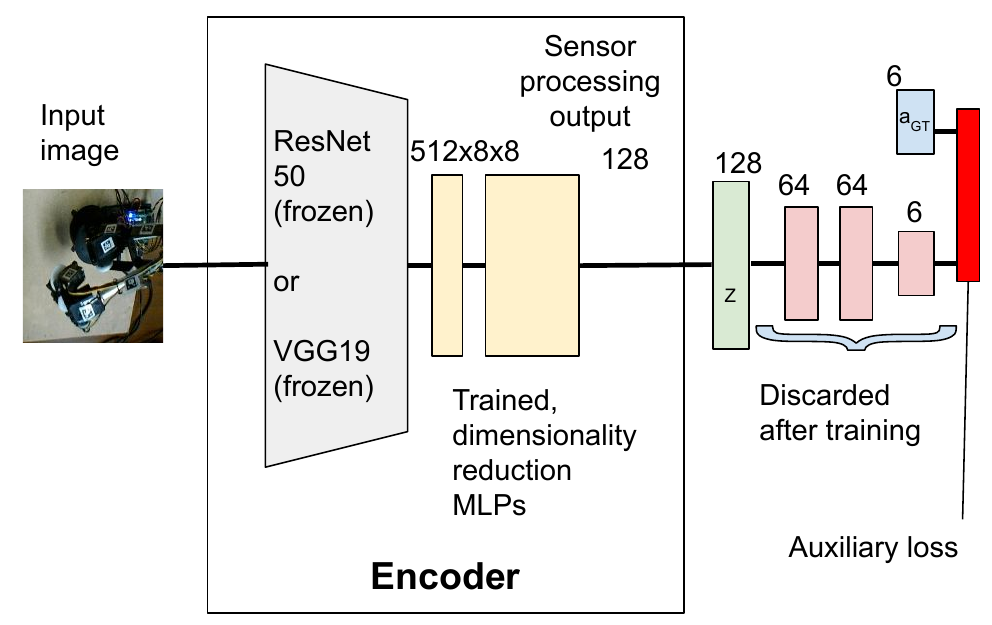}
        \caption{Encoder based on a pre-trained CNN fine-tuned for proprioception.}
        \label{fig:encoding-vgg19}
    \end{subfigure}
    
    \vspace{5mm}
    
    \begin{subfigure}[b]{0.35\linewidth}
        \centering
        \includegraphics[height=\myheight]{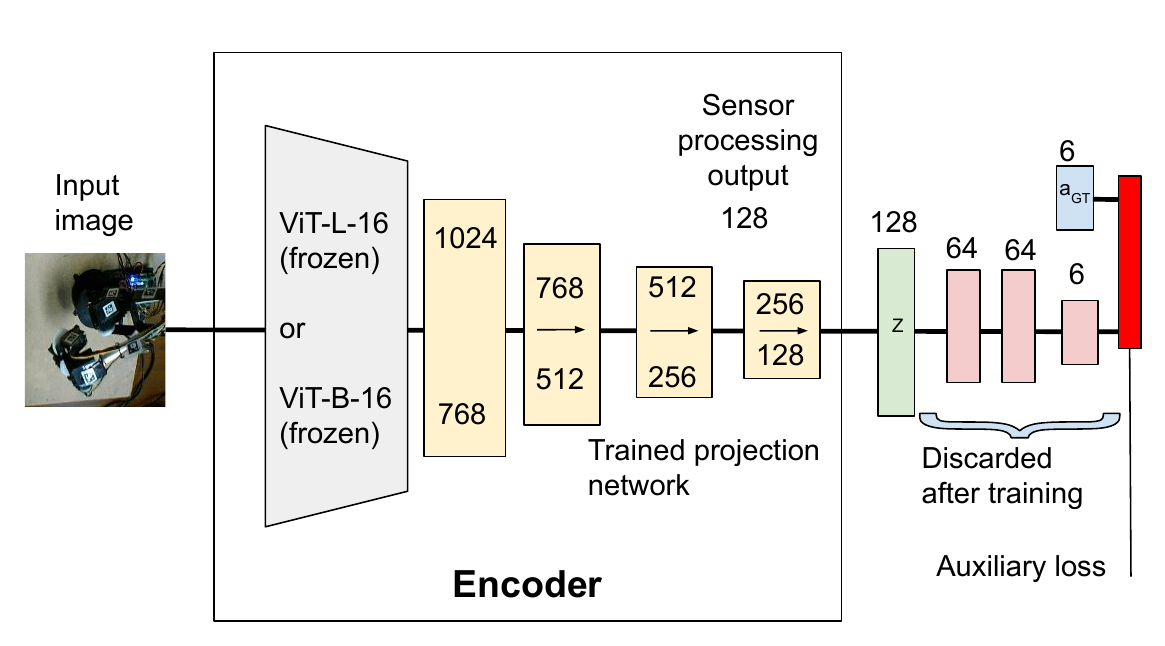}
        \caption{Encoder based on a pre-trained ViT fine-tuned for proprioception.}
        \label{fig:encoding-Vit}
    \end{subfigure}
    \hspace{5mm}
    \begin{subfigure}[b]{0.35\linewidth}
        \centering
        \includegraphics[height=\myheight]{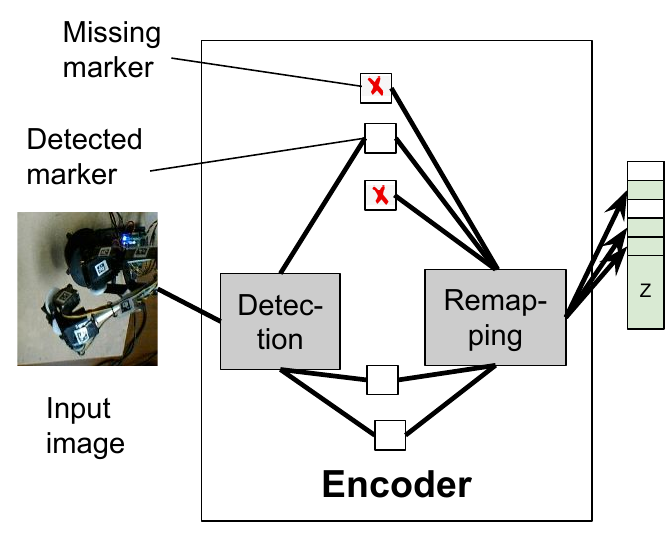}
        \caption{Encoder based on a bag of uncalibrated fiducial markers.}
        \label{fig:encoding-aruco}
    \end{subfigure}

    \caption{(a) Proprioception regression. From the observation $o$ the latent encoder creates the latent representation $\mathbf{z}$. The proprioception regressor creates an approximation $\hat{\mathbf{a}}$ of the robot configuration. (b-e) Four variations of encoders to obtain the proprioception-dedicated latent representation $\mathbf{z}_\textit{prop}$ (in green). For all encoders, the components outside the encoder block are supporting the surrogate losses and are discarded after training.}
\end{figure*}

\noindent {\bf Latent encoding generation using the encoder of a Convolutional VAE:} Convolutional Variational Autoencoders (Conv-VAE) are a variation of a VAE~\cite{kingma2013auto} that replaces the fully connected layers with convolutional layers, making it more suitable for the unsupervised learning of latent representations of images. Like the original VAE architecture, Conv-VAEs present the latent encoding in the form of a probability distribution described by a vector of mean values $\mu(o)$ and a vector of variances $\sigma^2(o)$. From these components, we retain only the $\mathbf{z} = \mu(o)$ value as the latent representation, discarding the sampling component and the decoder after training (see Figure~\ref{fig:encoding-convvae}). 

Conv-VAEs are trained using a pixel-wise reconstruction loss combined with a KL-divergence regularization term. Why would this representation be appropriate as a starting point for proprioception? We cannot expect an accurate reconstruction of a 256x256 image from a size-128 latent vector. We can, however, use the choice of the unsupervised training data to force the Conv-VAE to allocate more of its latent space to the representation of the robot pose. To achieve this, the training images were acquired by programming the robot to move to random poses against a {\em static} background. 

\noindent{\bf Latent representation based on a proprioception-tuned convolutional backbone:}
Convolutional backbones such as VGG, ResNet or EfficientNet, usually pretrained on ImageNet, are the workhorses of image processing tasks. They generate a feature representation $\mathbf{z}_\textit{feat}$ that can be used as a starting point for a regressor or classifier. A representative feature vector, for instance, for VGG-19 is the size of 8 $\times$ 8 $\times$ 512 = 32768. This is much larger than the size-128 $\mathbf{z}_\textit{prop}$ we are looking for. Intuitively, the CNN features can answer many questions about arbitrary images -- we, however, are only interested in answering questions related to robot pose about a particular robot. To reduce $\mathbf{z}_\textit{feat}$ to $\mathbf{z}_\textit{prop}$, we introduce an MLP-based reduction mechanism. It is not immediately obvious, however, how to train this reduction component. Our insight is that we temporarily extend this reductor to a full proprioception pipeline, and use a small amount of supervised training data to train it end-to-end. This would encourage the representation at the 128-size bottleneck to be a selection of features sufficient for proprioception -- even if the final training data and regressor is different. From this training process only the dimensionality reducer component is kept -- both the training data and the auxiliary loss component is discarded after fine-tuning (see Figure~\ref{fig:encoding-vgg19}). We repeated this process with VGG-19~\cite{Simonyan-2014-VGG} and ResNet-50~\cite{he2016deep} representations. 

\noindent{\bf Latent representation based on Vision Transformer backbone:} Vision Transformers (ViT)~\cite{Dosovitskiy-2020-ImageWorth} have emerged as powerful alternatives to convolutional neural networks for image processing tasks, demonstrating superior performance in interpretability and tasks that require considering long-range dependencies. Similar to convolutional backbones, these models are typically pretrained on large datasets like ImageNet and then adapted through transfer learning, in which a pretrained model serves as a feature extractor, and a small regressor or classifier is appended to handle a target task. Although the underlying model is very different, we face a similar problem as with convolutional backbones: the native output from popular pretrained models, such as ViT-B with its 768-dimensional  $\mathbf{z}_\textit{feat}$, ViT-L with 1024-dimensional $\mathbf{z}_\textit{feat}$, or ViT-H with 1280-dimensional  $\mathbf{z}_\textit{feat}$, is significantly larger than our desired compact representation $\mathbf{z}_\textit{prop}$. Our solution is similar to the one we used for CNN backbones: a projection architecture consisting of multiple fully-connected layers with batch normalization and dropout for regularization (Figure~\ref{fig:encoding-Vit}). 

\noindent{\bf Latent representation based on bags of uncalibrated fiducial markers:} Fiducial markers were a mainstay of object tracking, robotics, and AR/VR technologies before the deep learning revolution. They are designed to be localizable using very efficient algorithms, do not require training, and if calibrated, can provide a very good accuracy for 2D and 3D localization. The disadvantages of fiducial marker-based localization include the need to modify the environment by placing markers in highly visible and salient locations and on each object we aim to track. But the most practically problematic aspect is the need to calibrate the relative positions of the cameras and the tracked objects. 

Combining the marker technology with the abilities of deep learning-based regressors to extract information from incomplete, redundant and noisy data, we propose a hybrid solution based on {\em bags of uncalibrated fiducial markers} that can mitigate the challenges of the traditional marker deployment. We start by appending ArUco markers~\cite{garrido2014automatic}, square, binary-patterned images with a thick black border, at various locations on both sides of the robot arm. In contrast to the traditional deployment model, these markers will not be calibrated, there is no need for careful placement with the only requirement being sufficient marker coverage.

The latent encoding is simply a list of the normalized, uncalibrated detection values and a 0/1 visibility flag (see Figure~\ref{fig:encoding-aruco}). As the ArUco detection algorithm returns 8 values for a detection corresponding to the coordinates of the four corners of the marker, 10 markers mounted on the robot create a representation of size (8+1) $\times$ 10 = 90 which is zero-padded to 128 for uniformity with the other encodings. 

\section{Experimental results}
\label{sec:Experiments}

To test the ability of the proposed architecture to perform proprioception, we implemented nine models with two kinds of latent widths, covering the approaches described above: Conv-VAE based (128 and 256), proprioception fine-tuned VGG-19 (128 and 256), ResNet-50 (128 and 256), ViT-Base (128) and ViT-Large-based (256) and bags of uncalibrated fiducial markers-based (128). The regressor was individually trained for each encoding. In this section, we discuss the accuracy results obtained for the side camera in the setup from Figure~\ref{fig:RobotSetup}. Details about the data collection setup and the results using the front camera are provided in the Appendix.

Figure~\ref{fig:CompareAll} compares the average componentwise accuracy of the nine regressors. We note that the accuracy varies by component indicating the relative difficulty of estimating each configuration component: the heading appears to be the easiest, followed by the distance, height and wrist angle values. The hardest components to estimate are the wrist rotation, followed by the gripper's open/closed position. 

The second interesting observation relates to the size of the representations: we would expect the larger, size-256 representations to be more accurate, but this is only true for the convolutional VAE. The reason is likely because this representation has been trained using a ``generalist" autoencoding surrogate loss, thus it needs to use larger parts of the encoding for representations outside the robot itself. In contrast, for the proprioception-tuned VGG-19 model the smaller encoding generally outperforms the larger one, sometimes with a significant margin. The conclusion we can infer from this is that, appropriately tuned, the latent representation of size 128 is sufficient to convey the proprioception information to the regressor. 

Finally, different representations appear to have specific areas of strength. The ArUco-based model is the best on the gripper state and second-best at the wrist rotation, but consistently performs the worst in everything else. ViT-Large and VGG-19 128 are among the best in almost all components but not at gripper state detection and, in the case of ViT-Large, not at wrist rotation. Their representational pairs ViT-Base and VGG-19 256, perform much worse. The VAE-based representations appear to be competitive only at heading detection, while the ResNet-50-based representations only at gripper state detection. 

To study how the different models track the ground truth at individual timepoints, Figures~\ref{fig:Cam2-VAE} through ~\ref{fig:Cam2-Aruco} show the proprioception results on a random but continuous trajectory, with each figure having the ground truth drawn in black. For readability purposes, each figure has at most two tracks in addition to the ground truth. An ideal visual proprioception component would have its estimation line closely overlapping with the ground truth.

\renewcommand{\mywidth}{0.3\linewidth}
\renewcommand{\myheight}{6cm}
\begin{figure*}[t]
    \centering
    \begin{subfigure}[b]{\mywidth}
        \centering
        \includegraphics[height=\myheight]{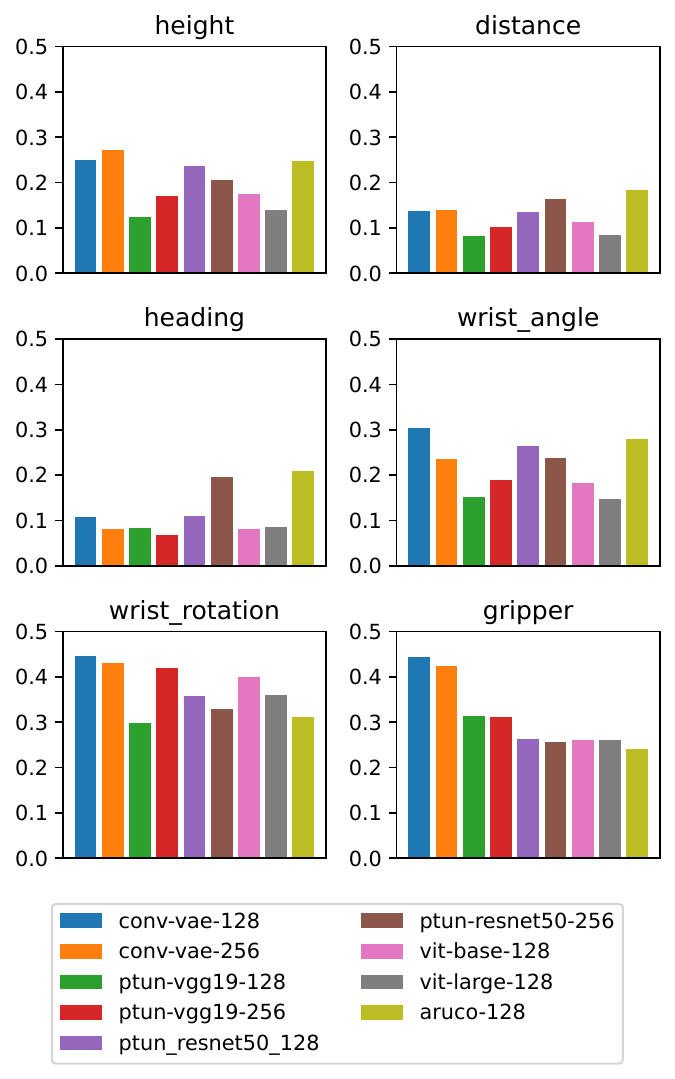}
        \caption{Mean squared error of the visual proprioception regression.}
        \label{fig:CompareAll}
    \end{subfigure}
    \hspace{5mm}
    \begin{subfigure}[b]{\mywidth}
        \centering
        \includegraphics[height=\myheight]{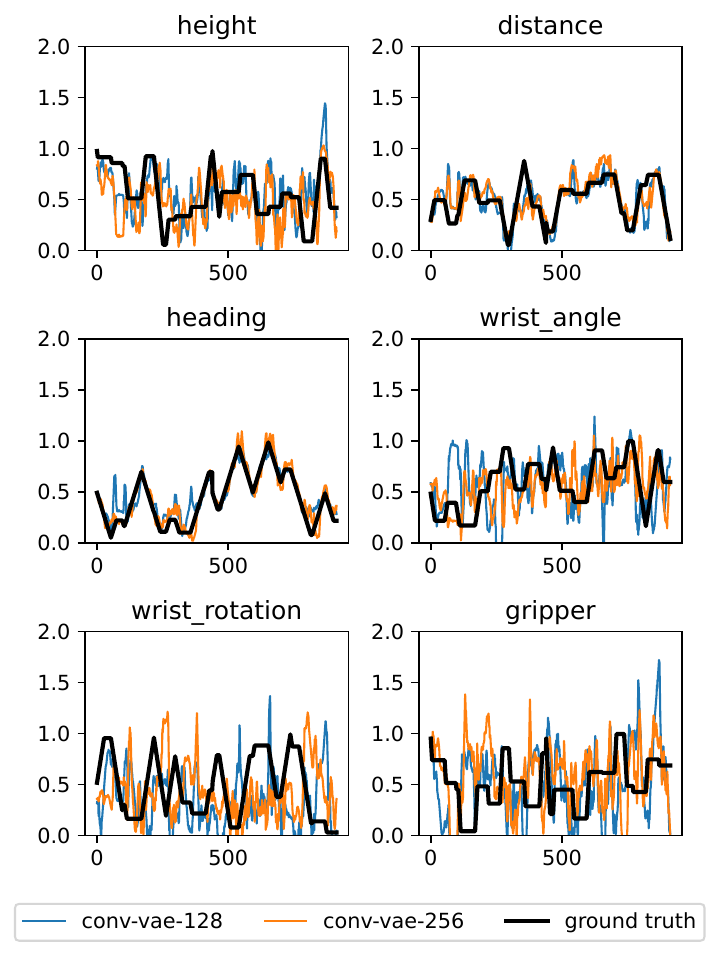}
        \caption{Convolutional variational autoencoder.\newline}
        \label{fig:Cam2-VAE}
    \end{subfigure}
    \hspace{5mm}
    \begin{subfigure}[b]{\mywidth}
        \centering
        \includegraphics[height=\myheight]{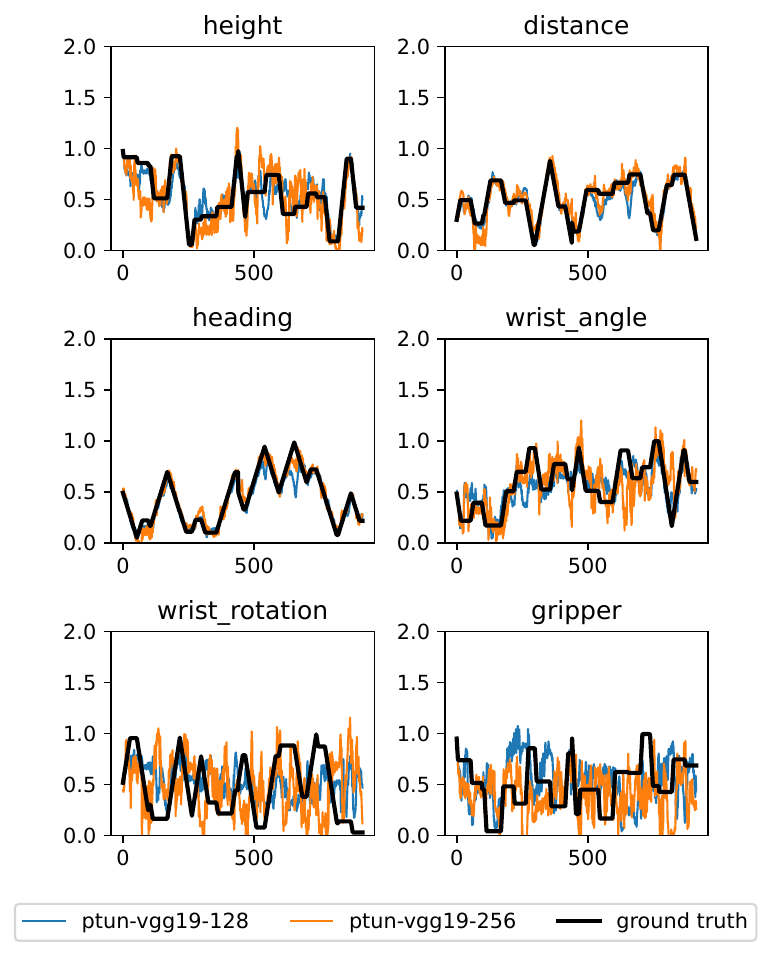}
        \caption{Proprioception-tuned VGG-19 encoding.}
        \label{fig:Cam2-VGG}
    \end{subfigure}
    
    \begin{subfigure}[b]{\mywidth}
        \centering
        \includegraphics[height=\myheight]{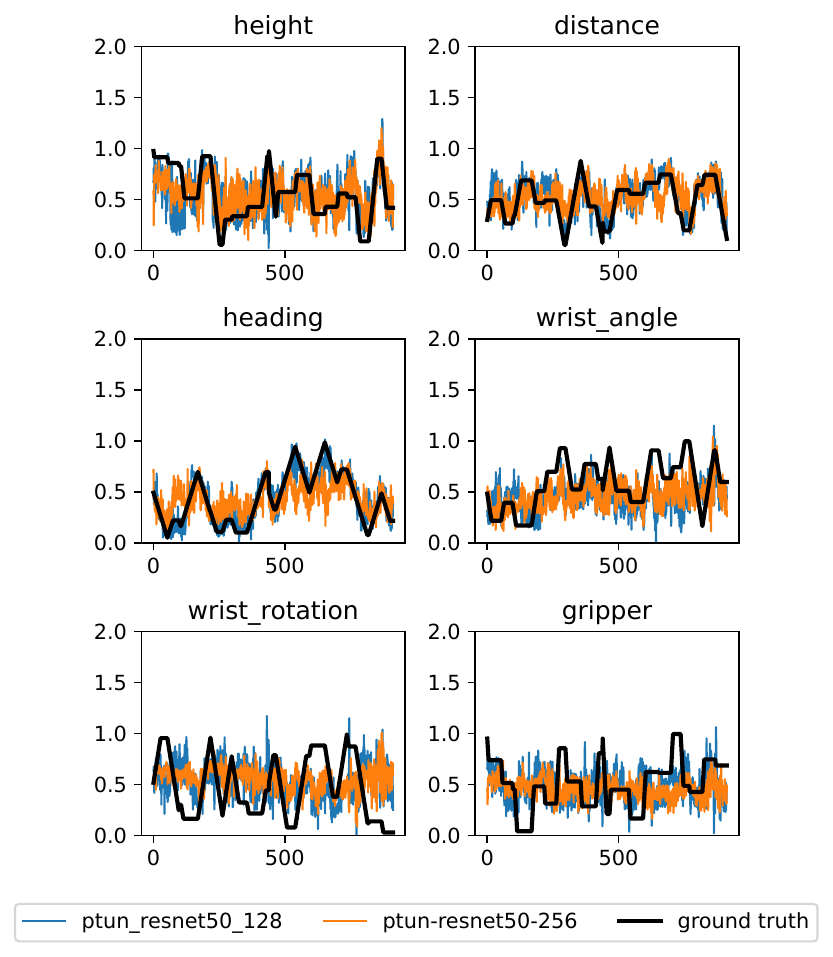}
        \caption{Proprioception-tuned ResNet-50 encoding.}
        \label{fig:Cam2-Resnet}
    \end{subfigure}
    \hspace{5mm}
    \begin{subfigure}[b]{\mywidth}
        \centering
        \includegraphics[height=\myheight]{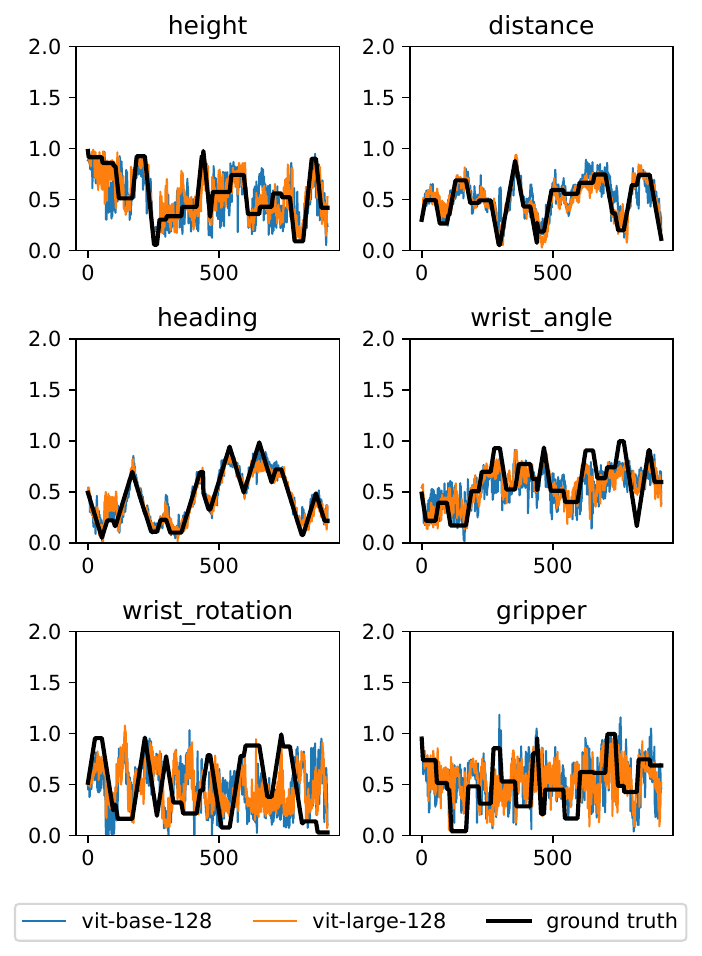}
        \caption{ViT-based latent encoding with ViT-Base and ViT-Large.}
        \label{fig:Cam2-Vit}
    \end{subfigure}
    \hspace{5mm}
    \begin{subfigure}[b]{\mywidth}
        \centering
        \includegraphics[height=\myheight]{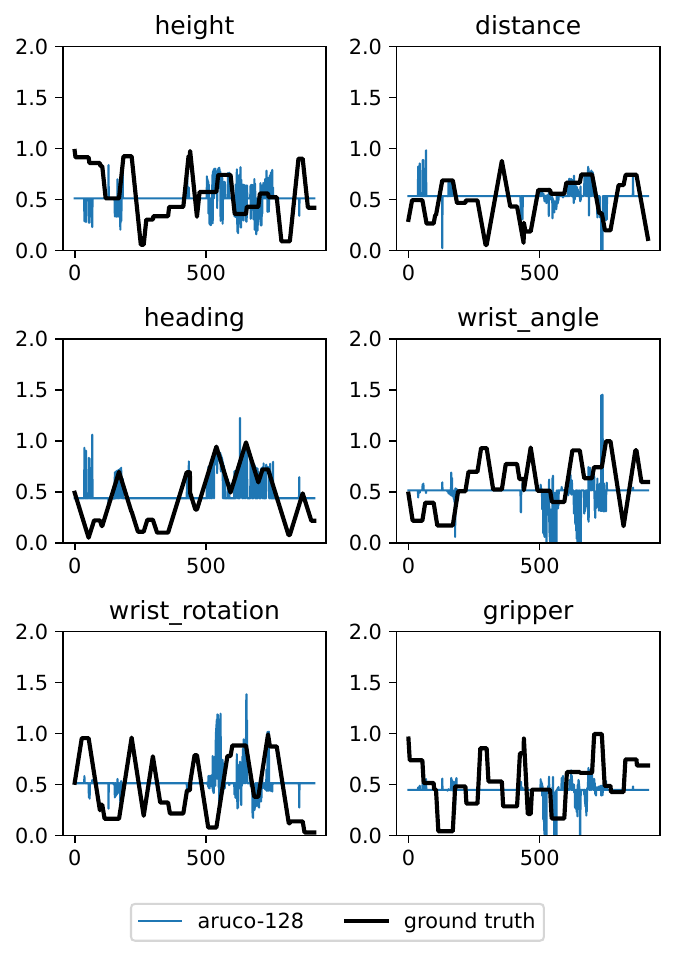}
        \caption{Bag of uncalibrated fiducial markers.\newline}
        \label{fig:Cam2-Aruco}
    \end{subfigure}
    \caption{Accuracy (a) and tracking (b-f) results based on observations from the side camera.}
\end{figure*}

Figure~\ref{fig:Cam2-VAE} shows the results for the convolutional VAE-based representations. As we have seen in the mean accuracy results, for this representation the larger embedding provides clear advantages for the heading and wrist angle detection - where the blue 128 line often produces large spikes. Although the 128- and 256-width VAEs were trained independently, the visual appearance of the tracking lines is quite similar, with a comparative stability of the track across successive frames. 

Figure~\ref{fig:Cam2-VGG} shows the results for the VGG-19 encoding which is among the best-performing representations overall. The estimated track is relatively smooth, without significant noise or spikes between observations. The only track that is relatively noisy is the one for the gripper status, which also provides the worst performance. Interestingly, the track of the size-256 latent is noisier, while having an overall similar appearance to the size-128 track.

Figure~\ref{fig:Cam2-Resnet} shows the results for the ResNet-50 encoding, one of the worst models from our collection. The tracks of both the size-128 and 256 representations are characterized by a very pronounced timestep-to-timestep noise, which is not present in the VGG-19 model nor the VAE models. This result is unexpected, because ResNet is often considered superior to VGG in most vision tasks. Possibly the low-level features pushed through the residual connections contribute to the instability of the final regression. 

Figure~\ref{fig:Cam2-Vit} shows the results for the ViT-based encoding. The tracks exhibit a noise similar to, but of lower amplitude than ResNet. The ViT-Large model is more accurate and avoids some of the spikes present in the ViT-Base model. 
 
Finally, Figure~\ref{fig:Cam2-Aruco} shows the result obtained with the embedding based on bags of uncalibrated ArUco markers. The fundamentally different nature of this encoding creates tracks with a very different overall look. Whenever a marker is detected, it will be localized precisely in the 2D observation -- but often only a small subset of markers is actually detected. Sometimes this is unavoidable, e.g. when the markers are on the other side of the robot, but sometimes it is due to conditions such as motion blur that changes from image to image. This on-and-off ``flickering" of detections makes the tracks very different from the other embeddings.

On components such as heading and distance, the ArUco-based model sometimes switches between almost perfect accuracy and the baseline of predicting the average pose. Interestingly, this phenomenon does not appear for the height value, which depends on the height of the gripper from the work surface. This might be due to the fact that there are no markers on the work surface.

\section{Conclusions}

In this paper, we investigated a range of latent representations supporting a fast, single-pass regression architecture for visual proprioception in inexpensive robots. Some of our findings were contrary to our initial expectations: the best tracking result was obtained by proprioception-tuned VGG-19, ResNet and ViT performed tracking with a strong noise between the timesteps, and the bag of uncalibrated fiducial markers representation led to a tracking alternating between high accuracy and random noise. 

Our results show promise for low computational overhead proprioception from observations that even inexpensive robot setups already have. This can improve control quality and extend the range of feasible tasks in unstructured environments. The results can also help by identifying parts of the robot's configuration that must have internal proprioception (in our setup, the grippers) and the ones that can be left to visual observation (in our case, determining the heading). 

As a future work direction, the results can be improved through techniques such as additional sensing data, temporal filtering, and more specialized vision algorithms.

%{\bf It has to be 8 pages up to here!}

% Remove for conference submission
\noindent {\bf Acknowledgments: } This work partly supported by the intramural research program of the U.S. Department of Agriculture, National Institute of Food and Agriculture via grant number 2024-67022-41788 and by NSF CPS grants \#1932300 and \#1931767.
% Any opinions, findings, conclusions, or recommendations expressed in this publication are those of the author(s) and should not be construed to represent any official USDA or U.S. Government determination or policy. 

\section{Limitations}
\label{sec:Limitations}

% All Submissions should include a Limitations section, explicitly describing limiting assumptions, failure modes, and other limitations of the results and experiments and how these might be addressed in the future.

Visual proprioception is strongly limited by the ability of the camera to capture relevant parts of the robot, and its ability to correctly interpret what it sees. Our experiments were performed in a benign environment, with constant, high quality lighting. In a practical deployment, additional occlusions will occur due to the manipulated objects, clutter and other distracting factors as well as artifacts of glare and low light. More general learning techniques, more powerful models and the use of multiple cameras can partially mitigate these problems. However, downstream robot control will need to take into consideration the limitations of this technology.

%============================================================================

% Appendices

\appendix
\renewcommand{\thesection}{\Alph{section}} % Labels sections as A, B, C...
\renewcommand{\thesubsection}{\alph{subsection}} % Labels subsections as a, b, c...

\section{Experimental settings}

The training and evaluation data were generated by programming the robot to perform random movements covering the configuration space. We recorded as ground truth the configuration achieved by the robot control in a free-space setting, and as input the recordings from two cameras. The {\em side camera} is situated to the right and above the robot and pointing down to the workspace at an angle of approximately 60 degrees. The {\em front camera} is situated directly in front of the robot, and it is also pointing down to the workspace at an angle of about 45 degrees. 

As the training data were recorded from continuous movement, the successive configurations form a smooth trajectory. These values were shuffled for training to increase the diversity of the batches of training data. For the evaluation data, we left the configurations in their original order for a cleaner visualization. However, no information was used from the temporal succession of frames, neither during training nor during inference. 

The collected training data were divided into four datasets, each with approximately 1000 entries. To ensure no information leakage between the datasets, no overlap was allowed between these sets. 
 
\begin{itemize}
\item {\bf Unsupervised training data:} used for the Conv-VAE-based encoder.
\item {\bf Supervised training data for fine-tuning:} used to train the proprioception-tuned encoder based on VGG-19, ResNet-50 and ViT backbones. 
\item {\bf Supervised training data for proprioception:} used to train the proprioception regressors for the different representations. To ensure the fairness of the comparisons, all nine configurations considered were trained with the same data. This dataset was split between training and validation data in an 80\% / 20\% ratio, used to determine the early stopping criterion. 
\item {\bf Test data:} used to evaluate the performance of the trained regressors.
\end{itemize}

\section{Experimental results with the front camera}

Figure~\ref{fig:AccuracyCam3} shows the results obtained when the experiments were repeated with a camera positioned in front of the robot. All the encoders and regressors were retrained from scratch. 

The overall results are very similar, with two notable differences. Due to the position of the camera, the detection of the heading value is much easier, leading to much better accuracy. Second, the accuracy of the encoding based on the bag of fiducial markers is much worse. This is due to the fact that most of the markers were positioned on the larger surfaces of the robot which are rarely visible from the front in most positions. 
 
\renewcommand{\mywidth}{0.45\linewidth}
\renewcommand{\myheight}{5.5cm}

\begin{figure}
    \centering
    \begin{subfigure}[b]{\mywidth}
        \centering
        \includegraphics[height=\myheight]{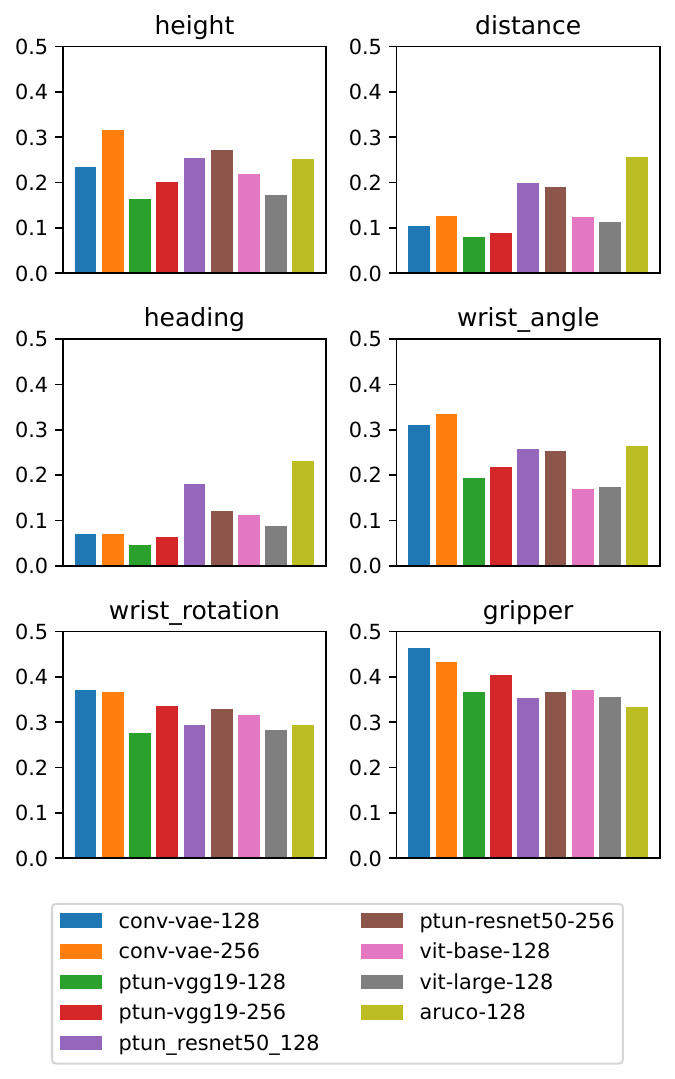}

        \caption{Mean squared error of the visual proprioception regression.}
        \label{fig:CompareAllCam3}
    \end{subfigure}
    \hspace{3mm}
    \begin{subfigure}[b]{\mywidth}
        \centering
        \includegraphics[height=\myheight]{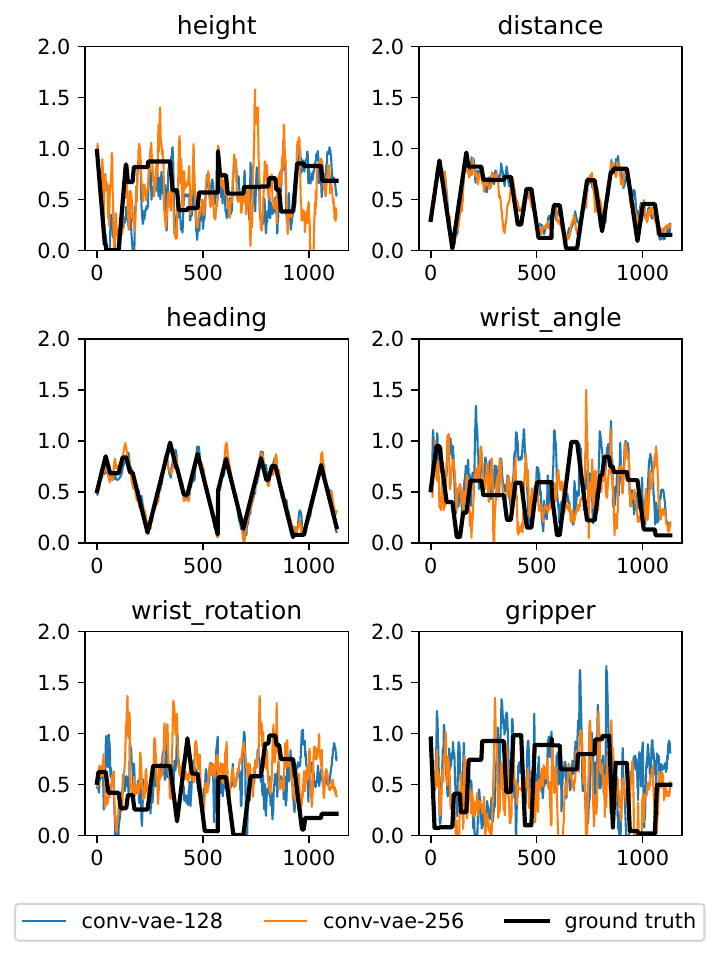}
        \caption{Convolutional variational autoencoder.\newline}
        \label{fig:Cam3-VAE}
    \end{subfigure}
    \caption{Accuracy (a) and tracking (b) results based on observations from the front camera.}
    \label{fig:AccuracyCam3}
\end{figure}

\section{Appendix 1}
\subsection{Technical Details and Hyperparameters}
This appendix provides technical details and hyperparameters used for training each model in the visual proprioception framework. Anonymous code\footnote{\url{https://github.com/Sahara-Sheik/Robotic-Proprioception}} and database\footnote{\url{https://doi.org/10.6084/m9.figshare.31743568}} links:

\vspace{-5mm}
\scriptsize
\begin{table}[H]
\centering
\caption{Architecture-Specific Details}
\label{tab:arch-details}
\vspace{-2mm}
\footnotesize
\begin{tabular}{|l|l|}
\hline
\textbf{Model} & \textbf{Details} \\
\hline
ResNet-50  & Reductor: 512; Proprioceptor: 64, 16 \\
ViT-Base   & $768\!\to\!384\!\to\!192\!\to\!128$ \\
ViT-Large  & $1024\!\to\!512\!\to\!256\!\to\!128$ \\
Conv-VAE   & Save period: 5 \\
\hline
\end{tabular}
\vspace{-6mm}
\end{table}
\vspace{-1mm}

\begin{table}[H]
\centering
\caption{Shared Training Configuration}
\label{tab:shared-config}
\vspace{-2mm}
\renewcommand{\arraystretch}{0.85}
\scriptsize
\begin{tabular}{|l|c|}
\hline
\textbf{Parameter} & \textbf{Value} \\
\hline
GPU & NVIDIA Tesla H100 80GB \\
Optimizer & Adam \\
Loss function & MSE \\
Training epochs & 300 \\
Output size (DoF) & 6 \\
Freeze backbone & True \\
Train / Val split & $\sim$80\% / 20\% \\
Dataset size per split & $\sim$1000 samples \\
\hline
\end{tabular}
\end{table}

\vspace{-6mm}

\begin{table}[H]
\centering
\caption{Model-Specific Hyperparameters}
\label{tab:model-specific}
\vspace{-2mm}
\renewcommand{\arraystretch}{0.85}
\scriptsize
\begin{tabular}{|l|c|c|c|c|}
\hline
\textbf{Model} & \textbf{Latent} & \textbf{LR} & \textbf{Batch} & \textbf{WD} \\
\hline
VGG-19       & 128/256 & $10^{-3}$ & 32 & -- \\
ResNet-50    & 128/256 & $10^{-3}$ & 32 & -- \\
ViT-Base     & 128     & $10^{-4}$ & 8  & 0.01 \\
ViT-Large    & 128     & $10^{-4}$ & 8  & 0.01 \\
Conv-VAE     & 128/256 & --        & -- & -- \\
\hline
\end{tabular}
\end{table}
\vspace{-6mm}
\begin{table}[H]
\centering
\caption{Per-joint MAE $\pm$ std and RMSE (normalized to $[0,1]$). \emph{Baseline} predicts the training-set mean.}
\vspace{-2mm}
\label{tab:results}
\renewcommand{\arraystretch}{0.85}
\setlength{\tabcolsep}{2pt}
\scriptsize
\begin{tabular}{llccc}
\toprule
Model & & height & distance & heading \\
\midrule
\multirow{2}{*}{conv-vae-128}
  & MAE & $.181 \pm .147$ & $.079 \pm .066$ & $.053 \pm .047$ \\
  & RMSE & .233 & .103 & .071 \\
\multirow{2}{*}{conv-vae-256}
  & MAE & $.239 \pm .206$ & $.096 \pm .083$ & $.054 \pm .045$ \\
  & RMSE & .316 & .127 & .070 \\
\multirow{2}{*}{vgg19-128}
  & MAE & $.121 \pm .109$ & $.061 \pm .051$ & $.037 \pm .028$ \\
  & RMSE & .163 & .079 & .047 \\
\multirow{2}{*}{vgg19-256}
  & MAE & $.161 \pm .120$ & $.070 \pm .054$ & $.052 \pm .038$ \\
  & RMSE & .201 & .089 & .064 \\
\multirow{2}{*}{resnet50-128}
  & MAE & $.200 \pm .156$ & $.164 \pm .112$ & $.134 \pm .120$ \\
  & RMSE & .254 & .199 & .180 \\
\multirow{2}{*}{resnet50-256}
  & MAE & $.217 \pm .163$ & $.153 \pm .115$ & $.088 \pm .082$ \\
  & RMSE & .272 & .191 & .121 \\
\multirow{2}{*}{vit-base-128}
  & MAE & $.169 \pm .138$ & $.098 \pm .077$ & $.078 \pm .081$ \\
  & RMSE & .218 & .125 & .112 \\
\multirow{2}{*}{vit-large-128}
  & MAE & $.130 \pm .113$ & $.088 \pm .071$ & $.064 \pm .060$ \\
  & RMSE & .172 & .113 & .088 \\
\multirow{2}{*}{aruco-128}
  & MAE & $.213 \pm .133$ & $.223 \pm .127$ & $.196 \pm .123$ \\
  & RMSE & .251 & .257 & .232 \\
\midrule
\multirow{2}{*}{Baseline}
  & MAE & $.169 \pm .146$ & $.223 \pm .127$ & $.197 \pm .122$ \\
  & RMSE & .223 & .257 & .232 \\
\bottomrule
\end{tabular}

\vspace{0mm}

\setlength{\tabcolsep}{2pt}
\scriptsize
\begin{tabular}{llccc}
\toprule
Model & & wrist\_ang & wrist\_rot & gripper \\
\midrule
\multirow{2}{*}{conv-vae-128}
  & MAE & $.249 \pm .186$ & $.300 \pm .219$ & $.390 \pm .250$ \\
  & RMSE & .310 & .372 & .463 \\
\multirow{2}{*}{conv-vae-256}
  & MAE & $.271 \pm .195$ & $.287 \pm .229$ & $.362 \pm .237$ \\
  & RMSE & .334 & .367 & .433 \\
\multirow{2}{*}{vgg19-128}
  & MAE & $.133 \pm .139$ & $.233 \pm .146$ & $.311 \pm .196$ \\
  & RMSE & .193 & .275 & .368 \\
\multirow{2}{*}{vgg19-256}
  & MAE & $.174 \pm .131$ & $.273 \pm .196$ & $.336 \pm .223$ \\
  & RMSE & .218 & .336 & .404 \\
\multirow{2}{*}{resnet50-128}
  & MAE & $.210 \pm .150$ & $.250 \pm .157$ & $.315 \pm .158$ \\
  & RMSE & .258 & .295 & .353 \\
\multirow{2}{*}{resnet50-256}
  & MAE & $.208 \pm .146$ & $.277 \pm .177$ & $.314 \pm .189$ \\
  & RMSE & .254 & .329 & .366 \\
\multirow{2}{*}{vit-base-128}
  & MAE & $.130 \pm .110$ & $.262 \pm .175$ & $.316 \pm .194$ \\
  & RMSE & .170 & .315 & .370 \\
\multirow{2}{*}{vit-large-128}
  & MAE & $.133 \pm .112$ & $.232 \pm .162$ & $.310 \pm .172$ \\
  & RMSE & .174 & .284 & .355 \\
\multirow{2}{*}{aruco-128}
  & MAE & $.216 \pm .154$ & $.246 \pm .159$ & $.294 \pm .156$ \\
  & RMSE & .265 & .293 & .333 \\
\midrule
\multirow{2}{*}{Baseline}
  & MAE & $.209 \pm .139$ & $.232 \pm .132$ & $.294 \pm .157$ \\
  & RMSE & .251 & .267 & .333 \\
\bottomrule
\end{tabular}
\end{table}

%%%%%%%%%%%%%%%%%%%%%%%%%%%%%%%%%%%%%%%%%%%%%%%%%%%%%%%%%%%%
%  I added this is this OK?
\bibliographystyle{IEEEtran}
\bibliography{main}
%\sah{does the bib meet their requirements?example provideded under ieeeconf/root.pdf}}
%%%%%%%%%%%%%%%%%%%%%%%%%%%%%%%%%%%%%%%%%%%%%%%%%%%%%%%%%%%%%%

\end{document}